\documentclass[conference]{IEEEtran}

\usepackage{cite}
\usepackage{amsmath,amssymb,amsfonts}
\usepackage{algorithmic}
\usepackage{algorithm}
\usepackage{graphicx}
\usepackage{textcomp}
\usepackage{xcolor}
\usepackage{booktabs}
\usepackage{multirow}
\usepackage{hyperref}
\usepackage{subcaption}
\usepackage{tikz}
\usepackage{pgfplots}
\pgfplotsset{compat=1.16}
\usetikzlibrary{shapes.geometric, arrows.meta, positioning, fit, calc, backgrounds, patterns}

\graphicspath{{figures/}}

\definecolor{revcolor}{RGB}{200,0,110}
\newcommand{\rev}[1]{#1}

\begin{document}

\title{FedChronos: Federated Fine-Tuning of Time-Series Foundation Models for Privacy-Preserving Commodity Price Forecasting}

\author{
\IEEEauthorblockN{
Amit Sharma\IEEEauthorrefmark{1},
Nitin Auluck\IEEEauthorrefmark{1},
and Akramul Azim\IEEEauthorrefmark{2}
}

\IEEEauthorblockA{
\IEEEauthorrefmark{1}Indian Institute of Technology Ropar, Punjab, India\\
Email: \{amit.21csz0028, nitin\}@iitrpr.ac.in
}

\IEEEauthorblockA{
\IEEEauthorrefmark{2}Ontario Tech University, Oshawa, Canada\\
Email: akramul.azim@ontariotechu.ca
}
}

\maketitle

\begin{abstract}
Time-series foundation models (TSFMs) such as Chronos have demonstrated strong forecasting capabilities across domains, yet adapting them to institutionally fragmented settings, where data cannot be centralized due to regulatory, competitive, or sovereignty constraints, remains unexplored. \rev{We introduce FedChronos, a framework for federated parameter-efficient fine-tuning of an already pre-trained TSFM, a setting that existing federated time-series work has not addressed, since prior methods either pre-train from scratch or align prototypes rather than adapt a fixed backbone.} Our approach applies Low-Rank Adaptation (LoRA) to the Chronos-T5 backbone and trains across distributed clients using FedAvg and FedProx, transmitting only lightweight adapter weights (384~KB per round, an 86$\times$ reduction over full-model exchange). We evaluate FedChronos on daily commodity prices from 15 Indian agricultural markets across 9 states, a naturally non-IID federated setting, and find that na\"ive LoRA fine-tuning overfits \rev{substantially} on small per-client datasets, dropping below zero-shot performance. \rev{We further observe that differential privacy (DP) noise can act as implicit regularization and counteract this overfitting: in our experiments the strongest configuration ($\varepsilon = 5$) reduces mean absolute percentage error (MAPE) by 31\% over zero-shot and 26\% over the best traditional baseline, while bounding each round's information leakage via per-round $(\varepsilon, \delta)$-differential privacy.} \rev{Because the model is compact and the updates are small, the approach also suits edge AI deployments where both the network link and the client device are constrained. Overall, our findings suggest that privacy and accuracy can be complementary rather than competing objectives in federated TSFM fine-tuning.}
\end{abstract}


\section{Introduction}

A potato farmer in Agra harvests 12 tonnes in March and has to decide what to do with the crop: sell now at the local mandi for Rs.~800/quintal, hold for two weeks and hope prices rise, or transport the harvest to the Delhi wholesale market 200~km away, where prices might reach Rs.~1,100, but transport costs may eat into the margin. Without reliable forecasts of how prices move across these markets, he relies on word of mouth and broker advice, information that is often days old and biased toward the broker's own interests. The same decision repeats across roughly 140 million farming households growing onions in Nashik, wheat in Amritsar, and rice in Kolkata, and the aggregate cost is large: information asymmetry alone is estimated to cost Indian farmers 15--20\% of their crop value annually~\cite{ladhar2023market_intelligence}.

The raw data needed to build better forecasts already exists. India operates over 7,000 regulated wholesale markets (mandis), each recording daily arrival quantities and modal prices for dozens of commodities. A model trained on this combined data could, in principle, learn cross-regional price dynamics: how a supply glut in Maharashtra's onion belt depresses prices in Kerala two weeks later, or how a cold snap in Uttar Pradesh ripples through potato markets from Agra to Kolkata. The obstacle is not data scarcity, but data fragmentation. State agricultural marketing boards operate under independent mandates, traders treat price-volume patterns as proprietary intelligence, and India's Digital Personal Data Protection Act (2023)~\cite{india_dpdp2023} adds compliance burdens to cross-entity data pooling. No single entity has access to all the data, and no institution is willing to hand over its share.

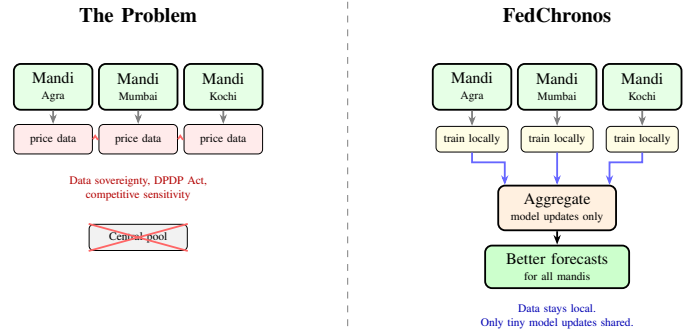
\begin{figure}[t]
\centering
\resizebox{\columnwidth}{!}{%
\begin{tikzpicture}[
    every node/.style={font=\scriptsize},
    mandi/.style={draw, rounded corners=3pt, minimum width=1.2cm, minimum height=0.7cm, fill=green!10, align=center, thick},
    silo/.style={draw, rounded corners=2pt, minimum width=1.2cm, minimum height=0.45cm, fill=red!8, align=center, font=\tiny},
    challenge/.style={draw, dashed, rounded corners=4pt, thick, red!50, inner sep=4pt},
    solution/.style={draw, rounded corners=4pt, thick, blue!50, fill=blue!5, inner sep=4pt},
    arr/.style={-{Stealth[length=1.5mm]}, thick},
    blocked/.style={thick, red!60, decorate, decoration={zigzag, segment length=4, amplitude=1.5}},
]

\node[font=\small\bfseries] at (-3.2, 3.3) {The Problem};

\node[mandi] (m1) at (-4.5, 2.2) {Mandi\\[-1pt]{\tiny Agra}};
\node[mandi] (m2) at (-3.2, 2.2) {Mandi\\[-1pt]{\tiny Mumbai}};
\node[mandi] (m3) at (-1.9, 2.2) {Mandi\\[-1pt]{\tiny Kochi}};

\node[silo] (s1) at (-4.5, 1.4) {price data};
\node[silo] (s2) at (-3.2, 1.4) {price data};
\node[silo] (s3) at (-1.9, 1.4) {price data};

\draw[arr, gray] (m1) -- (s1);
\draw[arr, gray] (m2) -- (s2);
\draw[arr, gray] (m3) -- (s3);

\draw[blocked] (s1.east) -- (s2.west);
\draw[blocked] (s2.east) -- (s3.west);

\node[font=\tiny, text=red!70!black, text width=2.8cm, align=center] at (-3.2, 0.65) {Data sovereignty, DPDP Act,\\competitive sensitivity};

\node[draw, rounded corners=2pt, fill=gray!10, minimum width=1.5cm, minimum height=0.4cm, font=\tiny, align=center] (pool) at (-3.2, -0.1) {Central pool};
\draw[thick, red!60] (pool.north west) -- (pool.south east);
\draw[thick, red!60] (pool.south west) -- (pool.north east);

\node[font=\small\bfseries] at (3.2, 3.3) {FedChronos};

\node[mandi] (fm1) at (1.9, 2.2) {Mandi\\[-1pt]{\tiny Agra}};
\node[mandi] (fm2) at (3.2, 2.2) {Mandi\\[-1pt]{\tiny Mumbai}};
\node[mandi] (fm3) at (4.5, 2.2) {Mandi\\[-1pt]{\tiny Kochi}};

\node[draw, rounded corners=2pt, fill=yellow!12, minimum width=1.1cm, minimum height=0.35cm, font=\tiny] (t1) at (1.9, 1.4) {train locally};
\node[draw, rounded corners=2pt, fill=yellow!12, minimum width=1.1cm, minimum height=0.35cm, font=\tiny] (t2) at (3.2, 1.4) {train locally};
\node[draw, rounded corners=2pt, fill=yellow!12, minimum width=1.1cm, minimum height=0.35cm, font=\tiny] (t3) at (4.5, 1.4) {train locally};

\draw[arr, gray] (fm1) -- (t1);
\draw[arr, gray] (fm2) -- (t2);
\draw[arr, gray] (fm3) -- (t3);

\node[draw, rounded corners=3pt, fill=orange!12, minimum width=1.8cm, minimum height=0.5cm, font=\scriptsize, align=center, thick] (srv) at (3.2, 0.35) {Aggregate\\[-1pt]{\tiny model updates only}};

\draw[arr, blue!60] (t1.south) -- ++(0,-0.15) -| (2.4, 0.7);
\draw[arr, blue!60] (t2.south) -- (3.2, 0.7);
\draw[arr, blue!60] (t3.south) -- ++(0,-0.15) -| (4.0, 0.7);

\node[draw, rounded corners=3pt, fill=green!20, minimum width=2.2cm, minimum height=0.45cm, font=\scriptsize, align=center, thick] (better) at (3.2, -0.55) {Better forecasts\\[-1pt]{\tiny for all mandis}};
\draw[arr] (srv) -- (better);

\node[font=\tiny, text=blue!70!black, text width=2.8cm, align=center] at (3.2, -1.3) {Data stays local.\\Only tiny model updates shared.};

\draw[gray, dashed] (0, 3.5) -- (0, -1.6);

\end{tikzpicture}%
}
\caption{The data sovereignty challenge in agricultural price forecasting. India's 7,000+ mandis each hold valuable price data, but cannot pool it due to regulatory (DPDP Act), competitive, and institutional barriers. FedChronos enables collaborative model improvement by sharing only lightweight model updates; raw data never leaves the mandi.}
\label{fig:motivation}
\end{figure}

Fig.~\ref{fig:motivation} illustrates this challenge, and it is not unique to Indian agriculture. Financial time series are siloed across banks, energy load data is partitioned across regional utilities, and healthcare signals are distributed across hospitals. In each domain the data exists in sufficient quantity for effective model training, but regulatory frameworks (GDPR~\cite{eu_gdpr2016}, HIPAA, sectoral data sovereignty laws), competitive dynamics, or plain institutional inertia prevent centralized aggregation. Collecting everything into one place and training a single model, the standard machine learning workflow, simply does not work here.

Foundation models have recently offered a promising starting point. Models such as Chronos~\cite{ansari2024chronos}, Lag-Llama~\cite{rasul2024lagllama}, TimesFM~\cite{das2024timesfm}, and TimeGPT~\cite{garza2024timegpt} re-purpose transformer architectures~\cite{vaswani2017attention} for temporal prediction, pre-training on large and diverse time-series corpora to learn general temporal patterns. The natural next step is domain-specific fine-tuning, adapting the pre-trained representations to the statistical properties of a target application. However, this step requires access to the target data, which brings us back to the fragmentation problem.

Federated learning (FL)~\cite{mcmahan2017fedavg} offers a way around this. Each institution trains on its own data and shares only model parameter updates, never raw observations, and a central server aggregates these updates into a global model that benefits from all participants' data, without any single party revealing theirs. Applying FL directly to foundation models, however, runs into a practical bottleneck: exchanging a full 8.4-million-parameter model every communication round across bandwidth-constrained rural networks is not feasible. Parameter-efficient fine-tuning through Low-Rank Adaptation (LoRA)~\cite{hu2022lora} resolves this by constraining updates to a small set of low-rank adapter matrices, under 2\% of the model, which cuts per-round transmission from megabytes to kilobytes. \rev{This same footprint, an 8.4M-parameter base model plus a 98,304-parameter adapter, fits comfortably on edge accelerators as well as server-class GPUs, which puts federated TSFM fine-tuning within reach of edge AI deployments where both the network link and the device itself are constrained.}

Federated LoRA has been explored for text-based large language models~\cite{gao2025fahqlora, wu2025survey_fedllm}, but no prior work has attempted federated parameter-efficient fine-tuning for time-series foundation models, in agriculture or any other domain. This gap has practical consequences: the institutions that need collaborative forecasting most, rural markets, regional utilities, small healthcare networks, have no path to benefit from pre-trained temporal models without surrendering their data.

This paper addresses that gap. We introduce \textbf{FedChronos}, a framework that applies LoRA-based federated fine-tuning to Chronos-T5-Tiny~\cite{ansari2024chronos}, a pre-trained time-series foundation model built on the T5 encoder-decoder architecture~\cite{raffel2020t5}. Each client, representing a regional agricultural market, fine-tunes only the LoRA adapter layers on its local price history, and transmits the adapter weights to a central server for aggregation via FedAvg~\cite{mcmahan2017fedavg} or FedProx~\cite{li2020fedprox}. The frozen base model parameters are never communicated. We further integrate client-level differential privacy~\cite{abadi2016dpsgd}, clipping and noising each round's adapter update to a per-round $(\varepsilon, \delta)$ target to bound the information leakage from any single update.

We evaluate FedChronos on real daily commodity price data from \rev{15 Indian mandis spanning 9 states, selected from the 464 that pass our data-quality filters,} a naturally non-IID setting where regional markets exhibit distinct price levels, seasonal patterns, and volatility profiles shaped by local geography, supply chains, and state-level agricultural policy. Our experiments show that differential privacy is not an optional safeguard layered on top of FedChronos, but a necessary part of it. Without DP, LoRA fine-tuning, whether centralized or federated, overfits on the small per-client datasets typical of this setting, and drops \emph{below} the Chronos zero-shot baseline (MAPE $\sim$102\%). \rev{With client-level DP enabled at $\varepsilon = 5$, the injected noise appears to act as implicit regularization and FedChronos reaches a MAPE of 69.78\% in this configuration, a 31\% improvement over zero-shot and 26\% over the strongest traditional baseline, while transmitting 86$\times$ less data per round than full-model exchange and bounding each round's information leakage via per-round $(\varepsilon, \delta)$-differential privacy.} Separately, FedProx reduces per-client accuracy variance compared to FedAvg in this heterogeneous setting, though the DP-driven accuracy gain is an order of magnitude larger than the gap between aggregation strategies. We therefore treat FedChronos as a DP-regularized framework by design, not as federated fine-tuning with privacy bolted on afterward.

Our contributions are as follows:
\begin{enumerate}
    \item \rev{We study federated parameter-efficient fine-tuning of an \emph{already pre-trained} time-series foundation model, a setting distinct from concurrent federated-TSFM work that pre-trains from scratch (FFTS~\cite{chen2025ffts}) or uses prototype alignment (FeDPM~\cite{deng2026fedpm}), and we identify an overfitting problem specific to it: na\"ive LoRA fine-tuning on small federated datasets degrades below zero-shot performance, a failure that holds across every unregularized method we test.}
    \item \rev{We observe that differential privacy noise can act as implicit regularization: our strongest configuration ($\varepsilon = 5$) yields a 31\% MAPE improvement over zero-shot and the best accuracy among the methods we evaluate. We report this as a promising effect and discuss its scope and current limits in Section~VI.}
    \item We compare FedAvg, FedProx, and local-only training under naturally non-IID conditions. FedProx reduces per-client variance, but the accuracy gain from DP regularization is an order of magnitude larger than the gap between aggregation strategies, so noise calibration matters more than the FedAvg-vs-FedProx choice in this setting.
    \item We show an 86$\times$ reduction in per-round communication cost via LoRA adapters (384~KB vs.\ 33~MB for full model exchange), making federated TSFM fine-tuning practical for bandwidth-constrained \rev{and edge} deployments.
\end{enumerate}

The remainder of this paper is organized as follows. Section~II surveys related work across time-series foundation models, federated learning with parameter-efficient fine-tuning, and agricultural applications. Section~III details the FedChronos methodology, including the Chronos tokenization scheme, LoRA adaptation, federated protocol, and differential privacy integration. Section~IV describes the experimental setup, dataset characteristics, and evaluation protocol. Section~V presents results and analysis. Section~VI discusses broader implications and limitations, and Section~VII concludes the paper.

\section{Related Work}

Our work sits at the intersection of three active research threads: time-series foundation models, federated learning with parameter-efficient fine-tuning, and machine learning for agricultural markets. We survey each in turn, highlighting the gap that FedChronos addresses.

\subsection{Time-Series Foundation Models}

The idea of pre-training large models on diverse time-series corpora and deploying them zero-shot on unseen forecasting tasks has gained rapid momentum since 2023. The approaches differ in architecture and training strategy, but share the premise that temporal patterns transfer across domains.

Chronos~\cite{ansari2024chronos} tokenizes real-valued time series into discrete bins via mean-scale uniform quantization and trains a T5 encoder-decoder~\cite{raffel2020t5} with cross-entropy loss, treating forecasting as sequence-to-sequence generation. It reaches competitive zero-shot performance across 42 benchmark datasets spanning energy, finance, weather, and retail demand. Lag-Llama~\cite{rasul2024lagllama} takes a decoder-only approach, conditioning on lag features for probabilistic forecasting. TimesFM~\cite{das2024timesfm} uses a patched-decoder architecture that processes fixed-length patches of the input series. TimeGPT~\cite{garza2024timegpt} takes a similar transformer-based approach, trained on a large proprietary corpus of time series and offered as a commercial forecasting API. TIME-LLM~\cite{jin2024timellm} reprograms frozen language models by aligning time-series patches with text prototypes, showing that linguistic pre-training captures temporal structure.

All of these models are designed and evaluated in centralized settings. To our knowledge, none has been adapted for federated training, a gap that matters once the target domain data is distributed across institutions that cannot, or will not pool their records.

\subsection{Federated Learning with Parameter-Efficient Fine-Tuning}

Federated learning~\cite{mcmahan2017fedavg} enables collaborative model training without centralizing data. In its canonical form (FedAvg), a server distributes a global model to participating clients, each client performs several epochs of local training, and the server aggregates the resulting model updates via weighted averaging. FedProx~\cite{li2020fedprox} extends this with a proximal regularization term that penalizes local model drift from the global parameters, which improves convergence when client data distributions are heterogeneous, the non-IID problem that affects most real-world FL deployments~\cite{zhao2018noniid}.

The cost of exchanging full model weights every FL round has motivated parameter-efficient fine-tuning (PEFT) within the federated pipeline. LoRA~\cite{hu2022lora} decomposes weight updates into products of low-rank matrices, training only about 1\% of parameters while preserving task performance, and QLoRA~\cite{dettmers2023qlora} extends this further by combining low-rank adapters with quantized base weights to fine-tune even larger models on a single GPU. In a federated setting, this means only the compact adapter weights need to travel each round rather than the entire model. \rev{This same compactness is what makes federated PEFT attractive at the edge, where memory, energy, and uplink bandwidth are all limited, and it underlies a growing body of work on edge-oriented federated adaptation.}

This combination has so far been explored only for text-based language models. FAH-QLoRA~\cite{gao2025fahqlora} combines heterogeneous quantization with LoRA for federated LLM fine-tuning across devices with varying compute budgets. Wu et al.~\cite{wu2025survey_fedllm} survey the emerging landscape of federated PEFT for language models and identify aggregation strategies and convergence properties as open problems. Edge-FIT~\cite{venkatesh2025edgefit} applies federated QLoRA instruction tuning in smart home settings.

The gap relevant to our work is that all of this federated PEFT research targets text-based models. Two concurrent works address federated learning of TSFMs from a different angle. FFTS~\cite{chen2025ffts} (AAAI'25) proposes federated \emph{pre-training} of time-series foundation models, training a model from scratch across heterogeneous client datasets with regularization to align shared knowledge. FeDPM~\cite{deng2026fedpm} employs discrete prototypical memories to align time-series data with language model representations in a federated setting. FedChronos is distinct from both: we address federated \emph{fine-tuning} of an already pre-trained TSFM using PEFT adapters, rather than training from scratch or using prototype alignment. This distinction matters practically, because organizations with limited data and bandwidth benefit most from adapting an existing pre-trained model with low-rank updates, rather than collaboratively training a new foundation model from scratch. \rev{While FFTS and FeDPM address federated learning of TSFMs from scratch or through prototype alignment, we are not aware of prior work that applies federated PEFT to adapt an \emph{already pre-trained} TSFM, which is the specific gap FedChronos targets.}

\subsection{Machine Learning for Agricultural Markets}

Federated learning has found agricultural applications mainly in crop disease classification~\cite{dembani2025agri_fl_review} and yield prediction~\cite{kumar2026fl_crop_market}. VLLFL~\cite{li2025vllfl} combines vision-language models with FL for agricultural object detection on edge devices. Separately, and without a federated learning component, AgriGen~\cite{kamduri2025agrigen} offers a prompt-tuned multilingual LLM-based question-answering system for farmer advisory, illustrating the broader push toward LLM-based agricultural tools that, like our work, still rely on centralized model access.

Agricultural \emph{price} forecasting has more direct economic impact on farmer livelihoods than yield estimation does~\cite{ladhar2023market_intelligence}, yet it has received surprisingly little attention in the FL literature. Existing price forecasting studies rely on centralized models~\cite{srivastava2025indian_price} that implicitly assume access to pooled data from multiple markets. In practice this assumption does not hold: mandis are run by different state bodies with no obligation, and often no willingness, to share data. Our work addresses this gap by bringing federated foundation model fine-tuning to the agricultural price domain, using real mandi data that carries the heterogeneity and institutional fragmentation of an actual deployment.

\section{Methodology}

This section presents the FedChronos framework in four parts: we first formulate the distributed forecasting problem, then describe how Chronos converts time-series forecasting into a language modeling task, explain how LoRA constrains the fine-tuning to a small parameter subspace, and finally detail the federated training protocol and its differential privacy extension.

\subsection{Problem Formulation}

Consider $K$ distributed clients (regional markets), each holding a local time-series dataset $\mathcal{D}_k = \{x_k^{(1)}, \ldots, x_k^{(n_k)}\}$ of daily commodity prices. The clients cannot share raw data due to regulatory, competitive, or institutional constraints. The goal is to collaboratively fine-tune a pre-trained time-series foundation model $f_\theta$ such that the global model minimizes the aggregate forecasting loss:
\begin{equation}
    \min_\theta \sum_{k=1}^{K} \frac{n_k}{N} \mathcal{L}_k(\theta), \quad N = \sum_{k=1}^{K} n_k
\end{equation}
Here, $\mathcal{L}_k(\theta) = \frac{1}{n_k} \sum_{i=1}^{n_k} \ell(f_\theta(x_k^{(i)}), y_k^{(i)})$ is the local empirical loss on client $k$, and $\ell$ denotes the per-sample forecasting loss.

\subsection{Chronos: Forecasting as Language Modeling}

Chronos~\cite{ansari2024chronos} recasts time-series forecasting as conditional token generation. This design choice is the main reason LoRA, originally developed for language models, transfers naturally to the time-series setting. Given a context window of $T$ real-valued observations, the model proceeds in three stages:

\textbf{Mean Scaling.} The input series is normalized by its mean absolute value, centering the distribution and making the quantization scheme robust to varying price scales across markets:
\begin{equation}
    \hat{x}_t = \frac{x_t}{\frac{1}{T}\sum_{i=1}^{T}|x_i| + \varepsilon}
\end{equation}

\textbf{Uniform Bin Quantization.} The scaled values are mapped to $B = 4{,}096$ discrete bins uniformly distributed between configurable limits $[\ell, u] = [-15, 15]$, producing integer token IDs. This turns the continuous forecasting problem into a classification problem over a fixed vocabulary, the same formulation used in language modeling.

\textbf{Encoder-Decoder Generation.} The token sequence is fed into a T5 encoder-decoder architecture~\cite{raffel2020t5}, trained to predict the next $H$ tokens (the forecast horizon) via cross-entropy loss. At inference, the decoder generates multiple stochastic forecast trajectories through temperature-scaled sampling, and point forecasts are computed as the median across samples.

The Chronos-T5-Tiny variant that we adopt uses $d_\text{model} = 256$, 4 encoder layers, 4 decoder layers, and 4 attention heads per layer, totaling 8.4M parameters. Its compact size makes it well-suited for federated settings where clients may have limited compute resources, while still capturing the temporal representations learned during pre-training on diverse time-series corpora.

\subsection{LoRA Adaptation of Chronos}

Fine-tuning all 8.4M parameters would erase the communication benefits of federated learning, since each client would then need to transmit the entire model every round. Instead, we apply LoRA~\cite{hu2022lora} to constrain updates to a low-dimensional subspace.

For each pre-trained weight matrix $\mathbf{W}_0 \in \mathbb{R}^{d \times d}$ in the attention mechanism, LoRA introduces a low-rank decomposition of the update:
\begin{equation}
    \mathbf{W} = \mathbf{W}_0 + \frac{\alpha}{r}\mathbf{B}\mathbf{A}, \quad \mathbf{B} \in \mathbb{R}^{d \times r}, \; \mathbf{A} \in \mathbb{R}^{r \times d}
\end{equation}
Here, $r \ll d$ is the rank, and $\alpha$ is a scaling factor that controls the magnitude of the adaptation relative to the pre-trained weights. The matrix $\mathbf{A}$ is initialized from a Gaussian distribution, $\mathbf{B}$ is initialized to zero, so the adaptation starts from the pre-trained model and gradually learns task-specific adjustments. Only $\mathbf{A}$ and $\mathbf{B}$ are trained; the original weights $\mathbf{W}_0$ remain frozen throughout.

We apply LoRA to the query ($\mathbf{W}_q$) and value ($\mathbf{W}_v$) projection matrices in every attention layer. With $r = 8$, $\alpha = 16$, and $d = 256$, each adapted matrix adds $2 \times 256 \times 8 = 4{,}096$ parameters. The Chronos-T5-Tiny architecture contains 24 attention projection matrices across its encoder self-attention (4 layers $\times$ $\{q, v\}$), decoder self-attention (4 layers $\times$ $\{q, v\}$), and decoder cross-attention (4 layers $\times$ $\{q, v\}$). The total trainable parameter count is therefore $24 \times 4{,}096 = 98{,}304$, representing just 1.16\% of the model. Table~\ref{tab:lora_config} summarizes this configuration.

\begin{table}[t]
\centering
\caption{LoRA configuration for Chronos-T5-Tiny. Only 1.16\% of parameters are trainable; the rest remain frozen.}
\label{tab:lora_config}
\begin{tabular}{lr}
\toprule
\textbf{Configuration} & \textbf{Value} \\
\midrule
Base model parameters & 8,394,496 \\
LoRA rank ($r$) & 8 \\
LoRA scaling ($\alpha$) & 16 \\
LoRA dropout & 0.05 \\
Target modules & $q$, $v$ projections \\
Adapted matrices & 24 \\
Trainable parameters & 98,304 (1.16\%) \\
Adapter size per round & 384 KB \\
\bottomrule
\end{tabular}
\end{table}

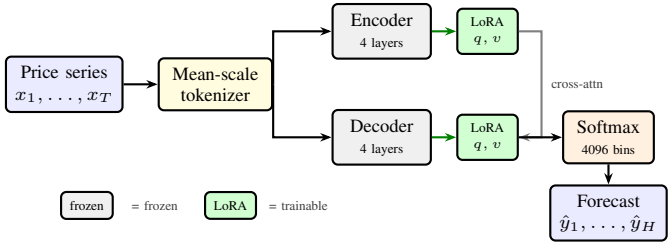
\begin{figure}[!t]
\centering
\begin{tikzpicture}[
    every node/.style={font=\scriptsize},
    box/.style={draw, rounded corners=2pt, minimum height=0.55cm, align=center, thick},
    frozen/.style={box, fill=gray!12, minimum width=1.3cm},
    lora/.style={box, fill=green!18, minimum width=1.3cm},
    io/.style={box, fill=blue!8, minimum width=1.5cm},
    arr/.style={-{Stealth[length=1.5mm]}, thick},
    lbl/.style={font=\tiny, text=gray!50!black},
]

\node[io] (input) at (-3.5, 0) {Price series\\$x_1,\ldots,x_T$};

\node[box, fill=yellow!15, minimum width=1.5cm] (tok) at (-1.5, 0) {Mean-scale\\tokenizer};
\draw[arr] (input) -- (tok);

\node[frozen] (enc) at (0.7, 0.7) {Encoder\\{\tiny 4 layers}};
\node[lora, minimum width=0.8cm, minimum height=0.4cm, font=\tiny] (lora_enc) at (2.1, 0.7) {LoRA\\$q,v$};
\draw[arr] (tok.east) |- (enc.west);
\draw[arr, green!50!black] (enc) -- (lora_enc);

\node[frozen] (dec) at (0.7, -0.7) {Decoder\\{\tiny 4 layers}};
\node[lora, minimum width=0.8cm, minimum height=0.4cm, font=\tiny] (lora_dec) at (2.1, -0.7) {LoRA\\$q,v$};
\draw[arr] (tok.east) |- (dec.west);
\draw[arr, green!50!black] (dec) -- (lora_dec);

\draw[arr, gray] (lora_enc.east) -- ++(0.3,0) |- (lora_dec.east) node[lbl, pos=0.25, right] {cross-attn};

\node[box, fill=orange!12, minimum width=1.2cm] (head) at (3.7, -0.7) {Softmax\\{\tiny 4096 bins}};
\draw[arr] (lora_dec) -- (head);

\node[io] (out) at (3.7, -1.7) {Forecast\\$\hat{y}_1,\ldots,\hat{y}_H$};
\draw[arr] (head) -- (out);

\node[frozen, minimum width=0.6cm, minimum height=0.3cm, font=\tiny] at (-3.2, -1.6) {frozen};
\node[lbl] at (-2.3, -1.6) {= frozen};
\node[lora, minimum width=0.6cm, minimum height=0.3cm, font=\tiny] at (-1.3, -1.6) {LoRA};
\node[lbl] at (-0.4, -1.6) {= trainable};

\end{tikzpicture}
\caption{Internal architecture of FedChronos at each client. A price series is tokenized via mean-scale uniform binning, passed through the frozen Chronos-T5 encoder-decoder, and decoded into forecast bins. Only the LoRA adapters (green) on the $q$ and $v$ projections are trainable; all other parameters remain frozen.}
\label{fig:architecture}
\end{figure}

\subsection{Federated Training Protocol}

\begin{figure}[t]
\centering
\includegraphics[width=\columnwidth]{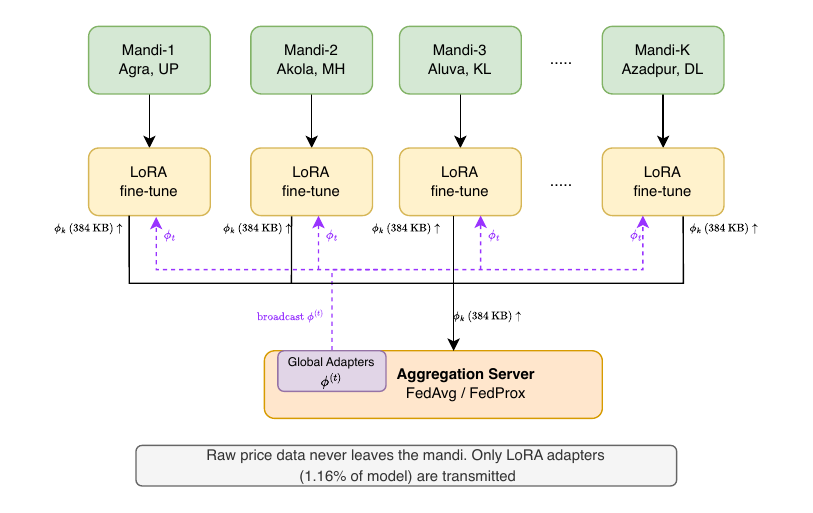}
\caption{FedChronos federated workflow. Each mandi fine-tunes LoRA adapters locally and uploads only the lightweight adapter weights (384~KB) to the server for aggregation. The server broadcasts the updated global adapters for the next round. Raw data never leaves the client.}
\label{fig:overview}
\end{figure}

With the LoRA parameterization in place, the federated protocol operates exclusively on the adapter parameters $\phi$ ($|\phi| = 98{,}304$), leaving the frozen base model untouched. We implement this protocol as a direct simulation loop, rather than through a dedicated FL framework such as Flower~\cite{beutel2020flower}, since our setting (a fixed set of 15 clients on a single GPU) does not require the process orchestration and device heterogeneity support that such frameworks are built for. Fig.~\ref{fig:architecture} illustrates the internal architecture at each client\rev{, where only the green LoRA adapters are trained and the rest stays frozen}; Fig.~\ref{fig:overview} shows the overall federated workflow\rev{, where only the 384~KB adapter crosses the network and raw data never leaves the mandi}; and Algorithm~\ref{alg:fedchronos} describes the procedure formally.

\begin{algorithm}[t]
\caption{FedChronos: Federated LoRA Fine-Tuning}
\label{alg:fedchronos}
\begin{algorithmic}[1]
\STATE \textbf{Server:} Initialize global LoRA parameters $\phi^{(0)}$
\FOR{each round $t = 1, \ldots, R$}
    \STATE Broadcast $\phi^{(t-1)}$ to all $K$ clients
    \FOR{each client $k = 1, \ldots, K$ \textbf{in parallel}}
        \STATE Load frozen Chronos-T5 base + $\phi^{(t-1)}$
        \STATE $\phi_k^{(t)} \leftarrow \textsc{LocalTrain}(k, \phi^{(t-1)}, E)$
        \IF{DP enabled}
            \STATE $\Delta_k \leftarrow \phi_k^{(t)} - \phi^{(t-1)}$
            \STATE $\Delta_k \leftarrow \textsc{ClipAndNoise}(\Delta_k, C, \sigma)$
            \STATE $\phi_k^{(t)} \leftarrow \phi^{(t-1)} + \Delta_k$
        \ENDIF
    \ENDFOR
    \STATE $\phi^{(t)} \leftarrow \sum_{k=1}^{K} \frac{n_k}{N} \phi_k^{(t)}$ \hfill \textit{// FedAvg aggregation}
\ENDFOR
\RETURN $\phi^{(R)}$
\end{algorithmic}
\end{algorithm}

Each client performs $E$ local epochs of training on its own price data before transmitting only the LoRA adapter parameters $\phi_k^{(t)}$ to the server. The server computes a weighted average, where each client's contribution is proportional to its local dataset size $n_k$. The key communication advantage is that only the adapter parameters (384~KB) are exchanged, rather than the full model (33~MB), leading to a reduction of 86$\times$.

\textbf{FedProx Extension.} In non-IID settings, where different clients have markedly different data distributions, FedAvg can suffer from client drift: local models diverge during training and their average may not represent any individual client well. FedProx~\cite{li2020fedprox} addresses this by adding a proximal regularization term to the local training objective:
\begin{equation}
    \mathcal{L}_k^{\text{prox}}(\phi) = \mathcal{L}_k(\phi) + \frac{\mu}{2} \|\phi - \phi^{(t-1)}\|^2
\end{equation}
Here, $\mu$ controls the strength of the penalty. This term discourages local adapters from straying too far from the current global consensus during each round. In our agricultural setting, where a mandi in Kerala may have fundamentally different price dynamics than one in Punjab, this regularization proves important for maintaining consistent performance across all participants.

\subsection{Threat Model and Privacy Guarantees}

We operate under the \emph{honest-but-curious} threat model standard in federated learning: the aggregation server and participating clients follow the protocol faithfully, but may attempt to infer private information from the model updates they observe. Concretely, the server receives LoRA adapter updates from each client and may try to reconstruct local price data or infer sensitive market intelligence (trading volumes, supplier relationships inferred from price correlations) from these updates.

To bound the information leakage from shared adapter weights, we integrate \emph{client-level} differential privacy (DP)~\cite{dwork2014algorithmic}. The protected unit is each client's entire local dataset: the DP guarantee ensures that an adversary observing a single round's adapter update cannot determine whether any given client participated, protecting the full training corpus of that market. This is client-level (dataset-level) DP, not example-level DP; example-level DP, which protects individual records, would require per-sample gradient clipping during local training and is not implemented here. Before transmitting its update $\Delta_k = \phi_k^{(t)} - \phi^{(t-1)}$, each client applies two operations:

\textbf{Norm Clipping.} The update is projected onto an $\ell_2$-ball of radius $C$:
\begin{equation}
    \bar{\Delta}_k = \Delta_k \cdot \min\left(1, \frac{C}{\|\Delta_k\|_2}\right)
\end{equation}
This ensures that no single data point in a client's local dataset can influence the shared update by more than a bounded amount, which is the precondition for the DP guarantee.

\textbf{Gaussian Noise Addition.} Calibrated noise is added to the clipped update:
\begin{equation}
    \tilde{\Delta}_k = \bar{\Delta}_k + \mathcal{N}(0, \sigma^2 C^2 \mathbf{I})
\end{equation}
Here, the noise multiplier $\sigma$ is calibrated to achieve a target privacy budget $(\varepsilon, \delta)$.

The privacy budget $\varepsilon$ quantifies the maximum information leakage: smaller $\varepsilon$ provides stronger privacy, but introduces more noise into the learning process. The parameter $\delta$ bounds the probability of a catastrophic privacy failure. We set $\delta = 10^{-5}$ (substantially smaller than $1/n$ for any client's dataset), and sweep $\varepsilon \in \{3, 5, 8\}$ in our experiments.

An important structural advantage of the LoRA-based approach is that the noise is distributed across only 98,304 parameters, rather than 8.4 million. For a fixed noise multiplier $\sigma$, the per-parameter signal-to-noise ratio is therefore higher than it would be under full-model DP-FL, making LoRA a natural fit for privacy-preserving federated fine-tuning.

We calibrate the noise multiplier $\sigma$ to achieve a target $(\varepsilon, \delta)$ guarantee for a \emph{single} round's update via the Gaussian mechanism, and apply this same calibration at every round. We do not perform formal privacy composition (e.g., via the moments accountant or R\'enyi DP~\cite{mironov2017renyi}) across the $R = 50$ communication rounds; the reported $\varepsilon \in \{3, 5, 8\}$ should therefore be interpreted as a per-round privacy parameter, rather than an end-to-end guarantee for the full training run. Composing the privacy loss across all 50 rounds would yield a substantially larger (weaker) cumulative $\varepsilon$ than the per-round value reported here. We leave rigorous end-to-end privacy accounting, and its interaction with the regularization effect we report, to future work.

\section{Experimental Setup}

\subsection{Dataset and Preprocessing}

We use daily commodity price data from the Agmarknet platform (Government of India), sourced via a publicly available Kaggle mirror\footnote{Dataset: \url{arjunyadav99/indian-agricultural-mandi-prices-20232025}}. The raw dataset spans June 2023 to June 2025, and contains 737,392 price records across 1,598 markets in 30 Indian states, covering five commodity categories: Potato, Onion, Wheat, Tomato, and Rice.

The preprocessing pipeline involves several steps needed to handle the messiness of real-world agricultural data. First, we standardize column names and parse dates (the raw data uses M/D/YYYY format). Multiple varieties of the same commodity are often recorded separately at a single mandi: a market might report prices for ``Potato (Local)'' and ``Potato (Red)'' on the same day, so we aggregate across varieties using the daily mean modal price. We then resample each market-commodity pair to daily frequency, interpolate gaps of up to 7 consecutive days (reflecting weekends and local holidays), and apply density filters to exclude series with more than 50\% missing values or fewer than 400 total observations. After filtering, 813 valid daily price series remain, covering Potato and Onion across 464 mandis. The temporal split allocates the first 70\% of each series to training, the next 15\% to validation, and the final 15\% to testing.

\subsection{Federated Client Partitioning}

Each mandi constitutes a natural FL client, mirroring the real-world institutional structure where each market operates independently under its state's agricultural marketing board. From the 464 available mandis, we select 15 for our experiments, spanning 9 Indian states with diverse climates, market scales, and trading volumes (Table~\ref{tab:clients}). The raw dataset spans June 2023 to June 2025 (approximately 735 daily observations per series after interpolation), yielding approximately 515 training, 110 validation, and 110 test observations per commodity series under the 70/15/15 temporal split.

\begin{table}[t]
\centering
\caption{Selected FL clients (mandis) for experiments, spanning 9 Indian states. Each client holds 2 commodity series (Potato and Onion); each series has approximately 515 training observations (1,030 per client across both commodities), drawn from the June 2023--June 2025 dataset with a 70/15/15 temporal split.}
\label{tab:clients}
\begin{tabular}{llll}
\toprule
\textbf{ID} & \textbf{Mandi} & \textbf{State} & \textbf{Region} \\
\midrule
1  & Achalda       & Uttar Pradesh  & North \\
5  & Ahmedabad     & Gujarat        & West \\
11 & Akola         & Maharashtra    & West \\
13 & Alipurduar    & West Bengal    & East \\
15 & Aluva         & Kerala         & South \\
16 & Alwar         & Rajasthan      & North \\
18 & Amritsar      & Punjab         & North \\
34 & Azadpur       & Delhi          & North \\
41 & Ballabhgarh   & Haryana        & North \\
42 & Ballia        & Uttar Pradesh  & North \\
45 & Banda         & Uttar Pradesh  & North \\
47 & Bangarmau     & Uttar Pradesh  & North \\
48 & Bankura Sadar & West Bengal    & East \\
49 & Bara Bazar    & West Bengal    & East \\
50 & Barabanki     & Uttar Pradesh  & North \\
\bottomrule
\end{tabular}
\end{table}

This partitioning is inherently non-IID along several dimensions. Geographically, the clients span from Kerala in the south to Punjab in the north, covering tropical, semi-arid, and subtropical climate zones that drive different seasonal price patterns. Economically, Azadpur (Delhi) is one of Asia's largest wholesale markets with high liquidity and tight bid-ask spreads, while Achalda (rural Uttar Pradesh) serves a primarily local catchment area with thinner trading volumes and more volatile prices. Commodity-wise, onion prices in Maharashtra's production belt behave very differently from onion prices in Kerala, which is a net-importing state where prices reflect transport costs and supply delays on top of the base production price.

\subsection{Baselines and Comparisons}

We compare seven methods spanning classical statistics, deep learning, and federated approaches:
\begin{itemize}
    \item \textbf{ARIMA}~\cite{box1976time}: Per-series auto-order ARIMA via \texttt{pmdarima}, fit independently on each test client's training data. This represents the classical statistical baseline.
    \item \textbf{LSTM}~\cite{hochreiter1997lstm}: Single-layer LSTM with hidden dimension $h=64$, trained on pooled data from all 15 clients. This represents the standard deep learning baseline with centralized data access.
    \item \textbf{Chronos Zero-Shot}: Pre-trained Chronos-T5-Tiny applied directly without any fine-tuning. This measures how well the foundation model's pre-trained knowledge transfers to Indian commodity prices.
    \item \textbf{Centralized LoRA}: LoRA fine-tuning on pooled data from all clients. This represents the privacy-violating upper bound, the best accuracy achievable if data sharing were unconstrained.
    \item \textbf{FedAvg LoRA}: Federated LoRA fine-tuning with standard weighted averaging~\cite{mcmahan2017fedavg}.
    \item \textbf{FedProx LoRA}: Federated LoRA with proximal regularization ($\mu = 0.01$)~\cite{li2020fedprox}.
    \item \textbf{Local-Only LoRA}: Each client fine-tunes its own LoRA adapter independently with no federated aggregation. This measures the benefit of federation.
\end{itemize}

\subsection{Training Configuration}

For centralized LoRA fine-tuning, we train with a learning rate of $10^{-3}$ (AdamW optimizer) for up to 5,000 steps. For federated experiments, we run 50 communication rounds with 3 local epochs per round and a batch size of 32. The context window is 256 time steps and the prediction horizon is 64 steps. DP experiments use a clipping norm of $C = 1.0$ with $\varepsilon \in \{3, 5, 8\}$ and $\delta = 10^{-5}$. The noise multiplier $\sigma$ for each target is calibrated via the Gaussian mechanism formula $\sigma = \sqrt{2 \ln(1.25/\delta)} / \varepsilon$ applied to a single round's update, yielding $\sigma \in \{1.615, 0.969, 0.606\}$ for $\varepsilon \in \{3, 5, 8\}$ respectively; as noted in Section~III, these are per-round values and do not account for composition across rounds. All experiments are conducted on an NVIDIA RTX 3090 GPU.

We evaluate using three standard forecasting metrics computed on the temporally held-out test data: Mean Absolute Error (MAE), Root Mean Squared Error (RMSE), and Mean Absolute Percentage Error (MAPE). MAE and RMSE are reported in the original price units (Indian Rupees per quintal), while MAPE provides a scale-invariant measure that allows comparison across commodities with different price levels. All reported metrics are macro-averaged across clients and commodities (i.e., each series is weighted equally regardless of length); per-client distributions are shown via box plots in Fig.~\ref{fig:fl_comparison}.

\section{Results}

We organize our experimental evaluation around four questions: (1) How do baselines and the pre-trained model perform on mandi data? (2) Does na\"ive fine-tuning improve over zero-shot? (3) How do FL algorithms compare, and what role does non-IID heterogeneity play? (4) How does differential privacy interact with fine-tuning? Does it simply cost accuracy, or is something more interesting going on?

\subsection{Baseline Performance}

Table~\ref{tab:baselines} presents baseline forecasting performance.

\begin{table}[t]
\centering
\caption{Baseline forecasting performance on mandi price data.}
\label{tab:baselines}
\begin{tabular}{lccc}
\toprule
\textbf{Method} & \textbf{MAE} & \textbf{RMSE} & \textbf{MAPE (\%)} \\
\midrule
ARIMA & 1597.87 & 1634.23 & 112.28 \\
LSTM (pooled) & 1342.47 & 1379.26 & 94.97 \\
Chronos Zero-Shot & 1413.46 & 1444.58 & 101.76 \\
\bottomrule
\end{tabular}
\end{table}

All baselines exhibit MAPE near or above 100\%, confirming the difficulty of Indian commodity price forecasting. The Chronos foundation model in zero-shot mode (MAPE 101.76\%) performs no better than predicting the series mean. This is expected: the Chronos pre-training corpus consists of Western financial, energy, and retail time series, which differ fundamentally from Indian agricultural prices driven by monsoon cycles, government MSP announcements, and regional supply shocks. The LSTM achieves the best baseline (MAPE 94.97\%) by learning cross-market patterns from pooled data, but requires centralized data access. \rev{With these baselines fixed, the question is whether fine-tuning the foundation model on local data beats its own zero-shot performance. It does not. Understanding why is where the more interesting result lies.}

\subsection{The Overfitting Challenge in TSFM Fine-Tuning}

Table~\ref{tab:main_results} shows a problem that, to our knowledge, has not been documented before: na\"ive LoRA fine-tuning of TSFMs on small federated datasets leads to severe overfitting, producing models that are worse than the zero-shot baseline.

\begin{table}[t]
\centering
\caption{Fine-tuning without regularization. All methods overfit, performing worse than the zero-shot baseline (MAPE 101.76\%).}
\label{tab:main_results}
\begin{tabular}{lccc}
\toprule
\textbf{Method} & \textbf{MAE} & \textbf{RMSE} & \textbf{MAPE (\%)} \\
\midrule
Chronos Zero-Shot & 1413.46 & 1444.58 & 101.76 \\
Centralized LoRA & 2001.99 & 2061.01 & 140.33 \\
FedAvg LoRA & 1986.51 & 2034.72 & 139.60 \\
FedProx LoRA & 1914.51 & 1963.28 & 134.82 \\
Local-Only LoRA & 1767.83 & 1812.45 & 123.79 \\
\bottomrule
\end{tabular}
\end{table}

With only 30 short price series (15 mandis $\times$ 2 commodities) and 98,304 trainable LoRA parameters, every unregularized fine-tuning method overfits: the model memorizes training patterns that do not generalize. Centralized LoRA, despite having access to all data, performs worst (MAPE 140.33\%), because pooling the heterogeneous price distributions from 9 states produces a noisy training signal that the model overfits to. Local-only training does best among the unregularized methods (MAPE 123.79\%), since each client's model can at least learn its own market's patterns, though it still overfits its limited local data.

\begin{figure}[t]
    \centering
    \includegraphics[width=\columnwidth]{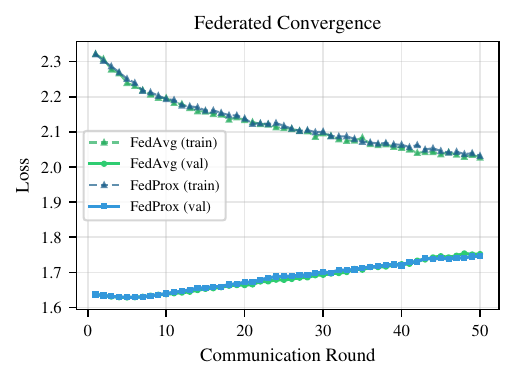}
    \caption{Validation loss during federated training. Both FedAvg and FedProx reach minimum validation loss within 5--7 rounds, and overfit thereafter. FedProx overfits \rev{marginally} more slowly due to its proximal regularization.}
    \label{fig:convergence}
\end{figure}

Fig.~\ref{fig:convergence} backs this up: validation loss reaches its minimum around round 5--7 and rises steadily after that, while training loss keeps falling, the classic overfitting signature. \rev{The effect holds across every unregularized method and both aggregation strategies: fine-tuning a TSFM on small, fragmented data hurts unless something holds the adaptation in check, an echo of similar findings for LoRA on small NLP datasets~\cite{hu2022lora}. Early stopping at the validation minimum is a natural complement to DP-based regularization here, and one we return to in Section VI.}

\subsection{FL Algorithm Comparison}

\begin{figure}[t]
    \centering
    \includegraphics[width=\columnwidth]{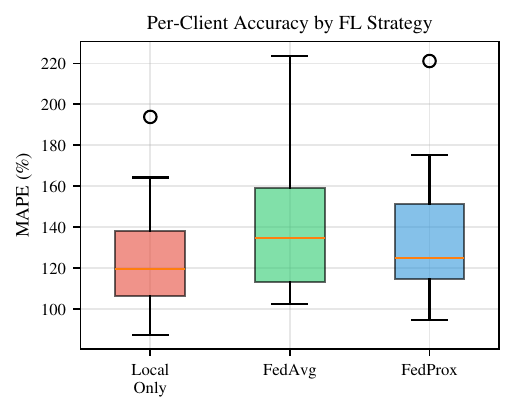}
    \caption{Per-client MAPE distribution. FedProx achieves lower variance than FedAvg, and local-only training yields the lowest median error among unregularized methods due to reduced cross-client interference.}
    \label{fig:fl_comparison}
\end{figure}

Fig.~\ref{fig:fl_comparison} compares per-client MAPE distributions. All unregularized methods overfit, but they do not all overfit the same way. \rev{Per-client consistency matters here: one market near 200\% MAPE is a real deployment failure regardless of the mean, and FedProx narrows that spread.}

FedProx (std=32.59) achieves lower per-client variance than FedAvg (std=32.96), since its proximal term keeps local models from drifting too far from the global consensus. Local-only training has the lowest aggregate MAPE (123.79\%) because each client avoids the interference that comes from averaging adapters trained on heterogeneous distributions. This points to a real difficulty for federated TSFM fine-tuning: regional commodity prices are non-IID enough that na\"ive parameter averaging can hurt more than it helps.

\subsection{DP as Implicit Regularization: The Key Finding}

\begin{figure}[t]
    \centering
    \includegraphics[width=0.85\columnwidth]{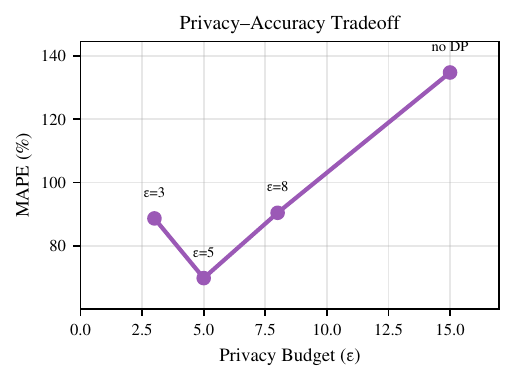}
    \caption{MAPE as a function of privacy budget $\varepsilon$. \rev{The curve suggests an apparent optimum near $\varepsilon = 5$}: too little noise (large $\varepsilon$) allows overfitting, while too much noise (small $\varepsilon$) overwhelms the learning signal.}
    \label{fig:dp}
\end{figure}

\rev{Neither aggregation strategy fixes the underlying overfitting, which raises the question of what does.} Table~\ref{tab:dp} and Fig.~\ref{fig:dp} present the main finding of this paper. \rev{Differential privacy noise appears to act as implicit regularization and turns models that were overfitting into the best performers we observe.}

\begin{table}[t]
\centering
\caption{DP-regularized federated fine-tuning. At $\varepsilon = 5$, DP noise prevents overfitting and achieves the best accuracy across all methods (\textbf{31\% lower MAPE than zero-shot, and 26\% lower than the best baseline, LSTM}).}
\label{tab:dp}
\begin{tabular}{lccc}
\toprule
\textbf{Privacy Budget ($\varepsilon$)} & \textbf{MAE} & \textbf{RMSE} & \textbf{MAPE (\%)} \\
\midrule
No DP ($\varepsilon = \infty$) & 1892.22 & 1941.82 & 134.73 \\
$\varepsilon = 8$ & 1180.14 & \textbf{1254.19} & 90.41 \\
$\varepsilon = 5$ & \textbf{988.86} & 1347.01 & \textbf{69.78} \\
$\varepsilon = 3$ & 1529.81 & 1914.17 & 88.64 \\
\bottomrule
\end{tabular}
\end{table}

Without DP, federated fine-tuning overfits to MAPE 134.73\%, performing worse than zero-shot. Adding DP noise at $\varepsilon = 5$ brings MAPE down to \textbf{69.78\%}, a 31\% improvement over Chronos zero-shot (101.76\%), and a 26\% improvement over the best baseline (LSTM at 94.97\%). Privacy budget and accuracy follow a U-shaped curve: too little noise ($\varepsilon = 8$, MAPE 90.41\%) gives insufficient regularization, while too much noise ($\varepsilon = 3$, MAPE 88.64\%) starts to overwhelm the learning signal. \rev{The best point we observe is at $\varepsilon = 5$, which pairs a reasonable per-round privacy budget with the strongest generalization in our sweep.}

\rev{Table~\ref{tab:dp} does carry one wrinkle that we want to flag directly. MAE and MAPE both favor $\varepsilon = 5$, but RMSE is lowest at $\varepsilon = 8$ (1254.19 versus 1347.01 at $\varepsilon = 5$). Because RMSE penalizes large errors more heavily than MAE, the most likely reading is that $\varepsilon = 5$ lowers the typical forecast error across most series while still leaving a few larger misses on high-price series, whereas $\varepsilon = 8$ distributes its error more evenly. We report all three metrics rather than only the most favorable one, and we interpret the $\varepsilon = 5$ result as the strongest point on the typical-error metrics rather than a clean win on every measure. Since these numbers come from a single training configuration per $\varepsilon$, we treat the exact location of the optimum as indicative rather than settled, and we return to this point in the limitations.}

\rev{A plausible mechanistic explanation is the following.} LoRA fine-tuning constrains updates to a 98,304-dimensional subspace, and on small datasets the optimizer can easily fit noise within that subspace, which is what leads to overfitting. The Gaussian noise injected by DP behaves much like other regularization techniques such as dropout or weight noise~\cite{neelakantan2015adding}: it keeps the adapter weights from settling into sharp minima that generalize poorly. This is consistent with theoretical work connecting differential privacy to generalization more broadly~\cite{dwork2015generalization}, where bounding the influence of any single training example also bounds how much a model can overfit to it. Because LoRA has far fewer parameters than full-model fine-tuning, the per-parameter noise magnitude at any given $\varepsilon$ is larger, which amplifies this regularization effect.

The practical takeaway is that in federated TSFM fine-tuning, privacy and accuracy \rev{need not be} in tension. DP noise, calibrated per round to an $(\varepsilon, \delta)$ target, bounds the information leakage of each shared update, while also acting as the regularization that makes fine-tuning work. This dual role depends on a specific combination of conditions: a pre-trained foundation model, low-rank adaptation, and small per-client datasets, conditions that are common in real federated deployments.

\subsection{Communication Efficiency}

Table~\ref{tab:comm} summarizes the communication overhead across LoRA rank configurations.

\begin{table}[t]
\centering
\caption{Communication cost per FL round. LoRA adapters reduce per-round transmission by 86$\times$ at rank $r = 8$, compared to full model exchange.}
\label{tab:comm}
\begin{tabular}{lrrrr}
\toprule
\textbf{Configuration} & \textbf{Params} & \textbf{KB/round} & \textbf{Compress.} \\
\midrule
Full model & 8,492,800 & 33,175 & 1$\times$ \\
LoRA $r=4$ & 49,152 & 192 & 172$\times$ \\
LoRA $r=8$ & 98,304 & 384 & 86$\times$ \\
LoRA $r=16$ & 196,608 & 768 & 44$\times$ \\
LoRA $r=32$ & 393,216 & 1,536 & 22$\times$ \\
\bottomrule
\end{tabular}
\end{table}

Over a complete training run (50 rounds, 15 clients, bidirectional communication), LoRA $r = 8$ requires 562~MB of total data transfer compared to 48.5~GB for full-model exchange. This 86$\times$ reduction matters beyond convenience: it determines whether the approach is feasible at all on the low-bandwidth, intermittent mobile connections typical of rural mandis. The LoRA-based approach is the difference between a system that works in practice and one that only works in simulation. \rev{The same footprint that keeps the uplink cheap also keeps on-device memory and compute modest, so running each client on an edge accelerator rather than a server-class GPU is realistic; we simulate all 15 clients on one GPU here, and testing this profile on real edge hardware such as Jetson-class or Raspberry Pi-class devices at the mandi is a follow-up we plan to pursue.}

\section{Discussion}

\subsection{Broader Implications}

Our central finding, that DP noise regularizes LoRA fine-tuning of pre-trained TSFMs on small federated datasets, has implications beyond agriculture. The conditions that produce this effect, a large pre-trained model adapted via low-rank updates, limited per-client data, and heterogeneous client distributions, arise naturally when energy utilities fine-tune load forecasting models or hospital networks adapt clinical time-series models. In these settings DP may work as a principled replacement for ad hoc regularization while also bounding the information leakage of each shared update.

The practical recipe is simple: freeze most parameters via LoRA, aggregate via FedAvg or FedProx, and calibrate the DP noise multiplier so it doubles as regularization. The 86$\times$ communication reduction from LoRA adapters keeps this feasible even over bandwidth-constrained rural networks, \rev{and the same small footprint makes the recipe a plausible building block for edge AI systems that need to adapt on-device without sending data to a central server.}

\subsection{Limitations}

\rev{Our study is a first step, and a few points frame how far its conclusions reach. The evaluation covers two commodity classes across 15 mandis in a simulated setting, so a production system would still have to contend with stragglers, dropped connections, Byzantine clients, and a move onto real edge hardware. Some of our choices are also deliberately conservative: we use the smallest Chronos variant, we fix a single small client cohort, and we tune $\varepsilon$ on this dataset, so the specific optimum near $\varepsilon = 5$ is best read as dataset-dependent guidance rather than a universal setting.}

\rev{The privacy side has room to grow in a similar way. We calibrate DP noise to a per-round $(\varepsilon, \delta)$ target and reuse that calibration each round, so the reported $\varepsilon$ is a per-round figure; a full end-to-end account via the moments accountant or R\'enyi DP would yield a larger cumulative value, and pairing that with secure aggregation is what a deployment-grade guarantee would require. Strengthening the empirical case is the other priority we see most clearly. The numbers here come from a single run per configuration, so confirming the U-shaped $\varepsilon$ curve across seeds, with confidence intervals, and adding a non-private control (early stopping at the validation minimum, or matched-magnitude noise without clipping) would pin down how much of the gain is specific to differential privacy rather than to regularization in general. These, together with a closer theoretical look at why DP helps here and a joint study of DP with FedProx, are the threads we intend to pursue in the journal extension of this work.}

\section{Conclusion}

\rev{We introduced FedChronos, a framework for federated parameter-efficient fine-tuning of a pre-trained time-series foundation model. We are not aware of prior work that adapts a fixed pre-trained TSFM through low-rank federated updates, which sets it apart from concurrent efforts that train such models from scratch.} Our experiments on Indian agricultural commodity prices uncover a real problem and an \rev{apparent} fix. Na\"ive LoRA fine-tuning, whether centralized or federated, overfits \rev{substantially} on small per-client datasets and ends up worse than the zero-shot baseline. Adding client-level differential privacy changes this outcome: at $\varepsilon = 5$, DP noise \rev{appears to act} as implicit regularization, cutting MAPE by 31\% over zero-shot, and 26\% over the best traditional baseline, while bounding each round's information leakage. This connection between privacy and accuracy \rev{is consistent with LoRA's small parameter space, where} the per-parameter noise magnitude regularizes the model without drowning out the learning signal.

\rev{Our results suggest that, in federated TSFM fine-tuning, privacy and generalization can be complementary rather than competing objectives, though confirming the strength and stability of this effect across seeds and against non-private regularization baselines is important future work.} The 86$\times$ communication reduction from LoRA adapters keeps the approach practical even over bandwidth-constrained \rev{and edge} networks. Future work would look at adaptive noise calibration, early stopping strategies, formal privacy composition across rounds, extension to larger TSFM variants, \rev{validation on real edge devices,} and deployment with real communication constraints. We plan to release code, pre-processing scripts, and data partition configurations upon acceptance to facilitate reproducibility.

\bibliographystyle{IEEEtran}
\bibliography{references}

@article{ansari2024chronos,
  title={{Chronos}: Learning the Language of Time Series},
  author={Ansari, Abdul Fatir and Stella, Lorenzo and Turkmen, Caner and Zhang, Xiyuan and Mercado, Pedro and Shen, Huibin and Shchur, Oleksandr and Rangapuram, Syama Sundar and Arango, Sebastian Pineda and Kapoor, Shubham and others},
  journal={arXiv preprint arXiv:2403.07815},
  year={2024}
}

@inproceedings{jin2024timellm,
  title={Time-{LLM}: Time Series Forecasting by Reprogramming Large Language Models},
  author={Jin, Ming and Wang, Shiyu and Ma, Lintao and Chu, Zhixuan and Zhang, James Y and Shi, Xiaoming and Chen, Pin-Yu and Liang, Yuxuan and Li, Yuan-Fang and Pan, Shirui and Wen, Qingsong},
  booktitle={International Conference on Learning Representations (ICLR)},
  year={2024}
}

@article{rasul2024lagllama,
  title={{Lag-Llama}: Towards Foundation Models for Probabilistic Time Series Forecasting},
  author={Rasul, Kashif and Ashok, Arjun and Williams, Andrew Robert and Ghonia, Hena and Bhagwatkar, Rishika and Khorasani, Arian and others},
  journal={arXiv preprint arXiv:2310.08278},
  year={2024}
}

@article{das2024timesfm,
  title={A Decoder-Only Foundation Model for Time-Series Forecasting},
  author={Das, Abhimanyu and Kong, Weihao and Sen, Rajat and Zhou, Yichen},
  journal={arXiv preprint arXiv:2310.10688},
  year={2024}
}

@article{garza2024timegpt,
  title={{TimeGPT}-1},
  author={Garza, Azul and Challu, Cristian and Mergenthaler-Canseco, Max},
  journal={arXiv preprint arXiv:2310.03589},
  year={2024}
}

@inproceedings{mcmahan2017fedavg,
  title={Communication-Efficient Learning of Deep Networks from Decentralized Data},
  author={McMahan, Brendan and Moore, Eider and Ramage, Daniel and Hampson, Seth and y Arcas, Blaise Aguera},
  booktitle={International Conference on Artificial Intelligence and Statistics (AISTATS)},
  year={2017}
}

@inproceedings{li2020fedprox,
  title={Federated Optimization in Heterogeneous Networks},
  author={Li, Tian and Sahu, Anit Kumar and Zaheer, Manzil and Sanjabi, Maziar and Talwalkar, Ameet and Smith, Virginia},
  booktitle={Conference on Machine Learning and Systems (MLSys)},
  year={2020}
}

@article{beutel2020flower,
  title={{Flower}: A Friendly Federated Learning Research Framework},
  author={Beutel, Daniel J and Topal, Taner and Mathur, Akhil and Qiu, Xinchi and Fernandez-Marques, Javier and Gao, Yan and Sani, Lorenzo and Li, Kwing Hei and Parcollet, Titouan and de Gusm{\~a}o, Pedro Porto Buarque and Lane, Nicholas D},
  journal={arXiv preprint arXiv:2007.14390},
  year={2020}
}

@inproceedings{hu2022lora,
  title={{LoRA}: Low-Rank Adaptation of Large Language Models},
  author={Hu, Edward J and Shen, Yelong and Wallis, Phillip and Allen-Zhu, Zeyuan and Li, Yuanzhi and Wang, Shean and Wang, Lu and Chen, Weizhu},
  booktitle={International Conference on Learning Representations (ICLR)},
  year={2022}
}

@inproceedings{dettmers2023qlora,
  title={{QLoRA}: Efficient Finetuning of Quantized Language Models},
  author={Dettmers, Tim and Pagnoni, Artidoro and Holtzman, Ari and Zettlemoyer, Luke},
  booktitle={Advances in Neural Information Processing Systems (NeurIPS)},
  year={2023}
}

@inproceedings{gao2025fahqlora,
  title={Federated Adaptive Fine-tuning of Large Language Models with Heterogeneous Quantization and {LoRA}},
  author={Gao, Zilong and Zhang, Zheming and Guo, Yulan and Gong, Yunchao},
  booktitle={IEEE INFOCOM},
  year={2025}
}

@article{wu2025survey_fedllm,
  title={A Survey on Federated Fine-tuning of Large Language Models},
  author={Wu, Yebo and Tian, Chunyu and Li, Junbo and Sun, Haozhao and Tam, Ka-Ho and Zhou, Zhicheng},
  journal={arXiv preprint arXiv:2503.12016},
  year={2025}
}

@article{li2025vllfl,
  title={{VLLFL}: A Vision-Language Model Based Lightweight Federated Learning Framework for Smart Agriculture},
  author={Li, Liang and Li, Jiaxing and Chen, Dong and Pu, Lingjuan and Yao, Haipeng and Huang, Yan},
  journal={arXiv preprint arXiv:2504.13365},
  year={2025}
}

@article{venkatesh2025edgefit,
  title={Edge-{FIT}: Federated Instruction Tuning of Quantized {LLMs} for Privacy-Preserving Smart Home Environments},
  author={Venkatesh, Vikram and Kamanuru, VR and Kumar, L and others},
  journal={arXiv preprint},
  year={2025}
}

@inproceedings{abadi2016dpsgd,
  title={Deep Learning with Differential Privacy},
  author={Abadi, Martin and Chu, Andy and Goodfellow, Ian and McMahan, H Brendan and Mironov, Ilya and Talwar, Kunal and Zhang, Li},
  booktitle={ACM Conference on Computer and Communications Security (CCS)},
  year={2016}
}

@article{dwork2014algorithmic,
  title={The Algorithmic Foundations of Differential Privacy},
  author={Dwork, Cynthia and Roth, Aaron},
  journal={Foundations and Trends in Theoretical Computer Science},
  volume={9},
  number={3--4},
  pages={211--407},
  year={2014}
}

@article{dembani2025agri_fl_review,
  title={Agricultural Data Privacy and Federated Learning: A Review of Challenges and Opportunities},
  author={Dembani, R and Karvelas, I and Akbar, NA and Rizou, S and others},
  journal={Computers and Electronics in Agriculture},
  year={2025},
  publisher={Elsevier}
}

@article{kumar2026fl_crop_market,
  title={Federated Learning-Based Approach for Crop Recommendation and Market Stability in Agriculture},
  author={Kumar, S and Maurya, T and Rai, M and others},
  journal={Federated Learning for Agriculture},
  year={2026},
  publisher={Wiley}
}

@inproceedings{srivastava2025indian_price,
  title={A Review Paper on the Study of Deep Learning and Machine Learning Models Used in Forecasting {Indian} Crop Prices},
  author={Srivastava, S and Dahiya, S},
  booktitle={International Conference on Data Analytics},
  year={2025},
  publisher={Springer}
}

@article{ladhar2023market_intelligence,
  title={{AI}-Based Market Intelligence Systems for Farmer Collectives: A Case Study from {India}},
  author={Ladhar, R and Sharma, S and Tangirala, S and Gupta, N and others},
  journal={ACM Journal on Computing and Sustainable Societies},
  year={2023}
}

@inproceedings{kamduri2025agrigen,
  title={{AgriGen}: A Prompt-Tuned, Multilingual {LLM}-Based {Q\&A} System for Smarter Agriculture},
  author={Kamduri, VRRKC and Gupta, P and El Kari, C},
  booktitle={Applied Imagery Pattern Recognition Workshop (AIPR)},
  year={2026},
  publisher={Springer}
}

@misc{india_dpdp2023,
  title={{The Digital Personal Data Protection Act}, 2023},
  author={{Government of India}},
  year={2023},
  howpublished={Ministry of Electronics and Information Technology}
}

@misc{eu_gdpr2016,
  title={Regulation ({EU}) 2016/679 — {General Data Protection Regulation}},
  author={{European Parliament and Council of the European Union}},
  year={2016}
}

@inproceedings{vaswani2017attention,
  title={Attention Is All You Need},
  author={Vaswani, Ashish and Shazeer, Noam and Parmar, Niki and Uszkoreit, Jakob and Jones, Llion and Gomez, Aidan N and Kaiser, Lukasz and Polosukhin, Illia},
  booktitle={Advances in Neural Information Processing Systems (NeurIPS)},
  year={2017}
}

@article{raffel2020t5,
  title={Exploring the Limits of Transfer Learning with a Unified Text-to-Text Transformer},
  author={Raffel, Colin and Shazeer, Noam and Roberts, Adam and Lee, Katherine and Narang, Sharan and Matena, Michael and Zhou, Yanqi and Li, Wei and Liu, Peter J},
  journal={Journal of Machine Learning Research},
  volume={21},
  number={140},
  pages={1--67},
  year={2020}
}

@article{neelakantan2015adding,
  title={Adding Gradient Noise Improves Learning for Very Deep Networks},
  author={Neelakantan, Arvind and Vilnis, Luke and Le, Quoc V. and Sutskever, Ilya and Kaiser, Lukasz and Karol, Kurach and Martens, James},
  journal={arXiv preprint arXiv:1511.06807},
  year={2015}
}

@article{hochreiter1997lstm,
  title={Long Short-Term Memory},
  author={Hochreiter, Sepp and Schmidhuber, J{\"u}rgen},
  journal={Neural Computation},
  volume={9},
  number={8},
  pages={1735--1780},
  year={1997},
  publisher={MIT Press}
}

@book{box1976time,
  title={Time Series Analysis: Forecasting and Control},
  author={Box, George E. P. and Jenkins, Gwilym M.},
  publisher={Holden-Day},
  address={San Francisco},
  year={1976}
}

@inproceedings{mironov2017renyi,
  title={R{\'e}nyi Differential Privacy},
  author={Mironov, Ilya},
  booktitle={IEEE 30th Computer Security Foundations Symposium (CSF)},
  pages={263--275},
  year={2017}
}

@article{zhao2018noniid,
  title={Federated Learning with Non-{IID} Data},
  author={Zhao, Yue and Li, Meng and Lai, Liangzhen and Suda, Naveen and Civin, Damon and Chandra, Vikas},
  journal={arXiv preprint arXiv:1806.00582},
  year={2018}
}

@inproceedings{dwork2015generalization,
  title={Generalization in Adaptive Data Analysis and Holdout Reuse},
  author={Dwork, Cynthia and Feldman, Vitaly and Hardt, Moritz and Pitassi, Toniann and Reingold, Omer and Roth, Aaron},
  booktitle={Advances in Neural Information Processing Systems (NeurIPS)},
  year={2015}
}

@inproceedings{chen2025ffts,
  title={Federated Foundation Models on Heterogeneous Time Series},
  author={Chen, Shengchao and Long, Guodong and Jiang, Jing and Zhang, Chengqi},
  booktitle={Proceedings of the AAAI Conference on Artificial Intelligence},
  year={2025}
}

@article{deng2026fedpm,
  title={Discrete Prototypical Memories for Federated Time Series Foundation Models},
  author={Deng, Liwei and Liu, Qingxiang and Niu, Xinhe and Chen, Shengchao and Sun, Sheng and Wu, Yuankai and Long, Guodong and Liang, Yuxuan},
  journal={arXiv preprint arXiv:2604.04475},
  year={2026}
}

\end{document}